\documentclass[letterpaper]{article} 
\usepackage{aaai25}  
\usepackage{times}  
\usepackage{helvet}  
\usepackage{courier}  
\usepackage[hyphens]{url}  
\usepackage{graphicx} 
\usepackage{natbib}  
\usepackage{caption} 
\usepackage{algorithm}

\usepackage{amssymb}
\usepackage{amsmath}
\usepackage{xspace}
\usepackage{graphicx}
\usepackage{url}

\usepackage{xcolor}
\usepackage{multirow}
\usepackage{newfloat}
\usepackage{listings}
\usepackage{tabularx}

\usepackage{xcolor}

\definecolor{bgcolor1}{HTML}{DAE8FC}
\definecolor{bgcolor2}{HTML}{D5E8D4}
\definecolor{bgcolor3}{HTML}{E1D5E7}

\DeclareCaptionStyle{ruled}{labelfont=normalfont,labelsep=colon,strut=off} 
\floatstyle{ruled}
\newfloat{listing}{tb}{lst}{}
\floatname{listing}{Listing}
\title{RATT: A Thought Structure for Coherent and Correct LLM Reasoning}
\author{
    Jinghan Zhang\textsuperscript{\rm 1},
    Xiting Wang\textsuperscript{\rm 2},
    Weijieying Ren\textsuperscript{\rm 3},
    Lu Jiang\textsuperscript{\rm 4},
    Dongjie Wang\textsuperscript{\rm 5},
    Kunpeng Liu\textsuperscript{\rm 1\thanks{Corresponding Author.}}
}
\affiliations{
    \textsuperscript{\rm 1}Portland State University\\
    \textsuperscript{\rm 2}Renmin University of China\\
    \textsuperscript{\rm 3}Pennsylvania State University\\ 
    \textsuperscript{\rm 4}Dalian Maritime University\\
    \textsuperscript{\rm 5}University of Kansas\\

    \{jinghanz,kunpeng\}@pdx.edu, xitingwang@ruc.edu.cn, wjr5337@psu.edu, jiangl761@dlmu.edu.cn, wangdongjie@ku.edu
}

\newcommand{\model}{RATT\xspace}
\newcommand{\modelbf}{\textbf{RATT}\xspace}

\usepackage{algpseudocode}
\usepackage{xcolor}
\usepackage{booktabs}

\definecolor{bgcolor1}{HTML}{DAE8FC}
\definecolor{bgcolor2}{HTML}{D5E8D4}
\definecolor{bgcolor3}{HTML}{E1D5E7}

\definecolor{txtcolor1}{HTML}{6C8EBF}
\makeatletter
\def\blfootnote{\gdef\@thefnmark{}\@footnotetext}
\makeatother

\begin{document}

\maketitle

\begin{abstract}
Large Language Models (LLMs) gain substantial reasoning and decision-making capabilities from thought structures. However, existing methods such as Tree of Thought and Retrieval Augmented Thoughts often fall short in complex tasks due to the limitations of insufficient local retrieval of factual knowledge and inadequate global selection of strategies. These limitations make it challenging for these methods to balance factual accuracy and comprehensive logical optimization effectively. To address these limitations, we introduce the Retrieval Augmented Thought Tree (\model), a novel thought structure that considers both overall logical soundness and factual correctness at each step of the thinking process. Specifically, at every point of a thought branch, \model performs planning and lookahead to explore and evaluate multiple potential reasoning steps, and integrate the fact-checking ability of Retrieval-Augmented Generation (RAG) with LLMs' ability to assess overall strategy. Through this combination of factual knowledge and strategic feasibility, the \model adjusts and integrates the thought tree structure to search for the most promising branches within the search space. This thought structure significantly enhances the model's coherence in logical inference and efficiency in decision-making, and thus increases the limit of the capacity of LLMs to generate reliable inferences and decisions based on thought structures. A broad range of experiments on different types of tasks showcases that the \model structure significantly outperforms existing methods in factual correctness and logical coherence.
\end{abstract}

%

\section{Introduction}

Large Language Models (LLMs) have shown impressive reasoning and decision-making capabilities in complex tasks like mathematical reasoning and creative writing by processing and generating thoughts based on token-level predictions~\cite{huang2022language,zhang2023planning}. However, this approach limits their ability to extend higher-level or multi-perspective reasoning~\cite{huang2022inner}. Research on LLM reasoning indicates that implementing structured thought processes significantly enhances their performance in inference and decision-making~\cite{yu2023towards}. These thought structures help LLMs to organize and generate their responses both contextually and hierarchically, as well as logically coherent reason across extended narratives and complex problem spaces.

Researchers have developed various thought structures for LLMs. Compared to the vanilla input-output (IO)~\cite{zhang2024instruct} prompting, where the model directly responds to the prompt, the basic idea of thought structures like Chain-of-Thought (CoT)~\cite{wei2022chain} and Self-consistency with CoT (CoT-SC)~\cite{wang2022self} is to guide the reasoning process by generating a sequential and coherent reasoning framework according to the given task. These methods usually involve generating a series of intermediate thought steps for analyzing and solving the initial question. They significantly improve LLM reasoning performance and the reasoning process transparency.

\begin{figure*}
    \centering
    \includegraphics[width=0.98\textwidth]{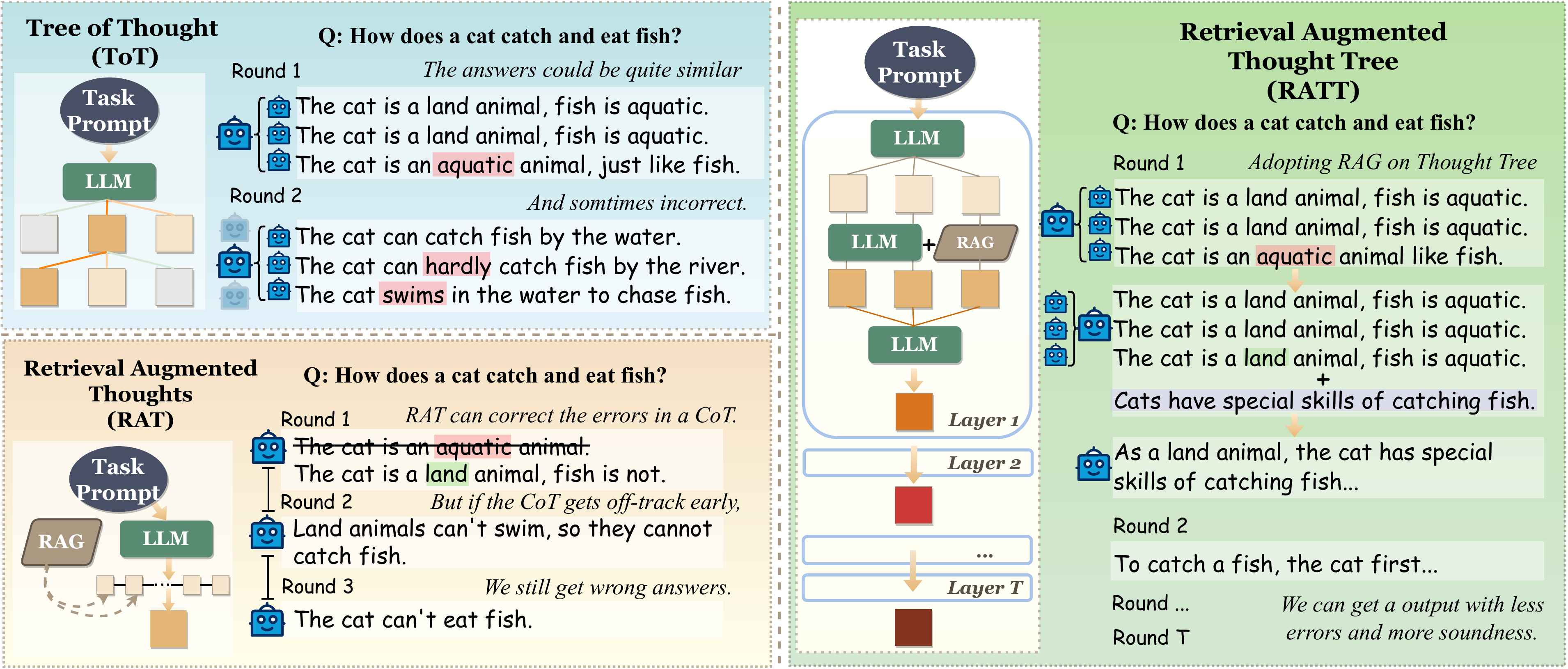}
    \caption{Comparison of LLM thought structures. With optimized tree structure and RAG, our method is capable of reducing factual and logical errors.}
    \label{fig:intro}
\end{figure*}

Despite the advancements in improving LLMs' reasoning capability, current thought structures still encounter a major challenge in balancing local factual accuracy and global strategy planning effectiveness. These limitations could restrict their applicability in complex and dynamic scenarios. 
For example, in Figure~\ref{fig:intro}, CoT and Tree of Thoughts (ToT)~\cite{yao2024tree}, which simulate coherent reasoning processes, lack an effective fact-checking mechanism. As a result, errors that occur in the early stage of reasoning can propagate through the entire thought chain and lead the conclusions to deviate far away from reality. Moreover, the global selection mechanism of the optimal branch of ToT is not effective and efficient enough, which could result in generating excessive texts and branches~\cite{ding2023everything}. While Retrieval Augmented Thoughts (RAT)~\cite{wang2024rat} combine CoT and Retrieval-Augmented Generation (RAG)~\cite{lewis2020retrieval} to perform fact checking, this method faces challenges in global strategy optimization, as it lacks systematic planning and lookahead to search for the optimal reasoning path in the search space and hence is limited in complex tasks where a broader perspective is necessary~\cite{zhao2024retrieval,renze2024self}.


\textbf{Our Targets.} To solve the aforementioned issues and enhance the reliability and accuracy of LLM reasoning, 
we aim to develop a method that seamlessly unifies local and global optimization.
At the \underline{\textbf{local}} level, the method should utilize external knowledge early and continuously to prevent and correct factual errors and thus avoid leading the model to incorrect search areas. If the factual errors are not identified early, they accumulate during subsequent generations and eventually lead to unreliable conclusions. 
At the \underline{\textbf{global}} level, the method should structure the reasoning process with planning and lookahead abilities. This allows for the identification of logically coherent and globally optimal reasoning paths even when there are frequent corrections of factual errors. Solely relying on comprehensive correction after the initial reasoning often leads to suboptimal solutions, as post-processing may not correct all factual errors or even introduce new errors due to delayed fact-checking.

\textbf{Our Approach.} To achieve these goals, we develop the Retrieval Augmented Thought Tree (RATT), a novel thought structure that simultaneously ensures both local factual correctness and global comprehensive logical soundness at each step of the reasoning process in one unified framework. In RATT, we perform planning and lookahead to multiple potential reasoning steps at each step of reasoning and integrate the fact-correction capability of RAG along with LLM's assessment of its overall strategy. Then, RATT integrates factual correctness and logical strategy feasibility to optimize the reasoning process and guide us toward the most promising branches within the search space. This approach greatly improves the logical coherence and inference reliability of LLM's decision-making and precise reasoning.

In summary, our contribution includes:
\begin{enumerate}
    \item We introduce a new thought structure called Retrieval Augmented Thought Tree (\model), which considers and improves comprehensive logical soundness and factual correctness at each step of the reasoning process. This structure significantly enhances the logical coherence and decision-making efficiency of LLMs in complex reasoning tasks.
    \item We develop a novel paradigm incorporating RAG into a tree-structured thought process. This paradigm conducts lookahead and general-to-detail fact-checking analyses at each node of the thought tree.
    \item We conduct a series of experiments to validate the effectiveness and robustness of our method (\model) across different tasks. Our experiments show that our method has clear advantages over existing methods.
\end{enumerate}

\section{Related Works}
\subsection{Thought Structures for LLMs}
Thought structures are a series of prompt engineering methods that guide models to generate more specific, accurate, and high-quality content~\cite{zhang2022automatic,minaee2024large}. Among these methods, Chain of Thought (CoT)~\cite{wei2022chain} is a milestone development that guides a reasoning process by generating a series of logically coherent intermediate steps of inference. Based on CoT, Wang et al. (2022)~\cite{wang2022self} introduce the Self-consistency with CoT (CoT-SC) method, which enhances reasoning accuracy and stability by independently generating multiple thought chains and selecting the most reliable answer. Furthermore, the ToT extends CoT by constructing a thought tree with multiple reasoning branches. The thought tree structure is capable of planning, looking ahead, and backtracking, thus providing a broad and global view of the solution space. These developments make multi-step thinking and reasoning similar to human cognitive processes.

\subsection{Retrieval-Augmented Generation for LLM Reasoning}
Retrieval-Augmented Generation (RAG)~\cite{lewis2020retrieval} for LLM reasoning has become a vital approach to enhance the quality of output. The LLM generates a response after retrieving information in an external library, which is a set of documents or knowledge related to the task, with a query relevant to the task~\cite{lewis2020retrieval,shuster2021retrieval}. This approach helps the LLM to produce responses that are more accurate, contextually relevant, and with fewer hallucinations~\cite{yang2024leandojo,wu2024retrieval}. Following RAG,~\cite{wang2024rat} developed the Retrieval-Augmented Thoughts (RAT) approach, which incorporates the RAG retrieval process into a reasoning chain-of-thought. The RAT approach first generates a reasoning chain and retrieves relevant information using LLM-generated queries based on the prompt and each reasoning step. Then RAT corrects and refines the reasoning chain step by step. The primary advantage of RAT is its ability to correct errors in the reasoning process, which is one notable step for enhancing the performance of LLM reasoning. However, as RAT only thinks and refines following one certain and complete path of thought, one potential problem of RAT is to fall into a local suboptimal solution in the search space.
\section{Methodology}

\begin{figure*}[h]
    \centering
    \includegraphics[width=0.85\textwidth]{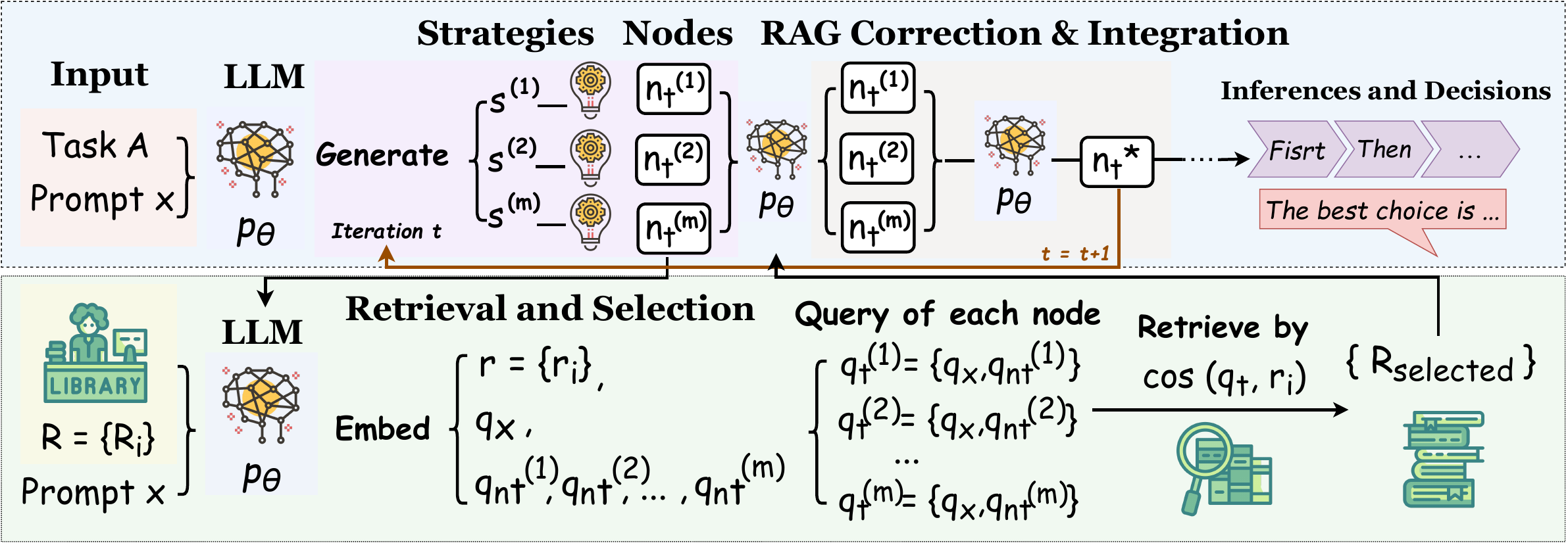}
    \caption{Framework of RATT. Given an input including a task, a prompt, and a set of documents, the LLM generates multiple thought nodes with different strategies and embeds the prompt and documents. Then, the model integrates the nodes into one optimized node. We use the embedding of the optimized node to retrieve and select relative documents. Finally, the LLM corrects possible mistakes and enriches the node with selected documents' information. This generation and optimization process iterates until reaching the maximum rounds of iteration.}
    \label{fig:method}
\end{figure*}

In this section, we introduce \underline{\textbf{R}}etrieval \underline{\textbf{A}}ugmented \underline{\textbf{T}}hought \underline{\textbf{T}}ree (\modelbf), an automated novel thought structure for language models that prioritizes both logical coherence and factual correctness. We aim to enhance the reliability and accuracy of LLM reasoning in two aspects: robust local fact-checking to prevent the accumulation of errors and global planning and lookahead to improve the logical coherence of reasoning paths. As shown in Figure~\ref{fig:method}, we implement this \model through several steps, including (1) \textbf{Thought Node Generation}, (2) \textbf{Retrieval and Selection}, and (3) \textbf{RAG Correction and Integration}.



\subsection{Problem Formulation}

To formalize our target and solution, first let us define the elements and functions. Given a task $\mathcal{A}$ and a pre-trained LLM $p_{\theta}$ with parameters $\theta$, we aim to enhance the performance of LLM on task $\mathcal{A}$ with performance metric $\mathbf{\epsilon} = \{ \epsilon_1, \epsilon _2, ..., \epsilon_j \}$. The task $\mathcal{A}$ can be any problem that can be solved with a logical sequence of thoughts or decisions. We have an input $x$ which is a language sequence and an output $y$. Our target is to construct the best thought tree structure $\mathcal{T}=\mathcal{T^{*}}$ capable of navigating the search space and optimizing the reasoning process with task $\mathcal{A}$ and prompt $x$:

\begin{equation}
  \mathcal{T}^* = \arg\max_{\mathcal{T}} \epsilon(y), \quad \text{where} \quad y = p_{\theta}(x | \mathcal{T}) .
\end{equation}
Here we denote a thought tree \(\mathcal{T} = (\mathcal{N}, \mathcal{E})\) to be a hierarchical tree structure consists of nodes and edges where each nodes represents a thought or decision, and an edge represents a logical flow between two nodes. Here \(\mathcal{N}\) is the set of nodes in the tree and \(\mathcal{E} \subseteq \mathcal{N} \times \mathcal{N} \) is the set of directed edges connecting the nodes.

\subsection{Thought Node Generation}

Our first step is to generate a thought tree that balances both exploration and exploitation adeptly and effectively. Given an prompt \( x \) with task information, the LLM \( p_{\theta} \) first embeds the prompt: 
\begin{equation}
    q_x = p_{\theta}(\text{embed}(x)).
\end{equation}

Then the LLM generates initial thoughts utilizing multiple logical strategies \( s^{(m)} \) to explore possible solutions in the search space of task $\mathcal{A}$ from a broad range of perspectives and logical approaches. Each initial thought represents a distinct logical viewpoint, and these thoughts collectively cover various possible or potential directions and solutions to tasks \( \mathcal{A}\). Specifically, in the \(t\)-th iteration, we define the initial thoughts to be the initial nodes of the tree: $\mathbf{n}_t = \{ n_t^{(1)}, n_t^{(2)},\ldots, n_t^{(m)}\}$. In this way, the thought tree takes every step with a broad and diverse range of \underline{\textit{logical possibilities}}. We denote a reasoning branch \(\mathcal{B}\) of the tree as a sequence of connected nodes starting from the root node and extending to any leaf node. Each branch \(\mathcal{B}\) represents a feasible solution for task \(\mathcal{A}\), and each node in the branch sequence is connected by an edge \(\mathcal{E}\) from its parent node, formulated as:

\begin{equation}
    \mathcal{B} = \{n_{root}, \ldots, n_{leaf}\},  (n_i \xrightarrow{p_\theta} n_{i+1}) \in \mathcal{E} .
\end{equation}
After generating initial nodes, the thought tree employs a planning and lookahead strategy, which is essential for dynamically evaluating the potential of branches. For each node \( n_t^{m} \), the LLM evaluates the quality of the immediate next steps as well as simulates the future states and consequences of the node's choice. The simulation provides insights into the likely outcomes to the model. This lookahead strategy helps the model to \underline{\textit{plan several moves ahead}}, refine and adjust generation strategies, and keep aligning the generation direction with the overall goal of the task.

\subsection{Retrieval and Selection}
Given a Library with $I$ candidate documents $\mathcal{R}:= \{ R_i \}_{i=1}^I $, the LLM embedes each document as:
\begin{equation}
    r_i = p_{\theta}(\text{embed}(R_i)) \in \mathbb{R}^K ,
\end{equation}
where $K$ is the dimension of the embedding and $\text{embed}(\cdot)$ is the embedding process. We then form a query $Q_t^{m}$ of each $n_t^{m}$ and the input $x$:

\begin{equation}
    q_t^{m} = \text{embed} \{ Q_t^{m} \} =\{ q_x, q_{n_t^{m}} \} \in \mathbb{R}^K, \text{where}
\end{equation}
\begin{equation}
    q_x = p_{\theta}(\text{embed}(x)), \text{and}
\end{equation}
\begin{equation}    
    q_{n_t^{m}} = p_{\theta}(\text{embed}(n_t^{m})).
\end{equation}
We then retrieve the whole Library to find the top-$k$ most relevant documents. Here we evaluate the relevance of each document $R_i$ to the query $Q_t^{m}$ by the cosine similarity of the embeddings:
\begin{equation}
    \text{sim}(q_t^{m}, r_i) = \frac{q_t^{m} \cdot r_i}{\|q_t^{m}\| \|r_i\|} .
\end{equation}

These selected documents \( \mathcal{R}_{\text{selected}} \) are then adept to enhance the information content of the initial nodes and \underline{\textit{support the fact-checking}}. Analogizing human learning and reasoning processes that go top to bottom from general understanding to specific details, we adopt a similar strategy in the retrieval process. We denote all nodes in the same round of generation to be in the same layer \(L\). Specifically, at the top layers \(L_1, L_2, ..., L_{l_1}, l_1 \in [1,l)\), we utilize a \textit{broad and high-level} retrieval strategy to gather basic concepts and background information; At the middle layers \(L_{l_1+1}, ..., L_{l_2}, l_2 \in [l_1+1,l)\), we utilize a \textit{targeted} retrieval strategy to acquire deeper information; At the leaf layers \(L_{l_2+1}, ..., L_{l}, l_2 \in [l_1+1,l)\), we utilize a \textit{detailed and specific} retrieval strategy to correct specific errors and enrich the details.




\subsection{RAG Correction and Integration}
We then correct and integrate the initial nodes with relevant documents $\mathcal{R_{\text{selected}}}$. In this process, the LLM \( p_{\theta} \) analyzes the documents \( \mathcal{R}_{\text{selected}} \) and extracts the most critical and supplementary information relevant to \(n_t^{1}, n_t^{2}, \ldots, n_t^{m}\). The information includes factual details, background explanations, logical support, and counter-examples. The model $p_{\theta}$ then integrates them into a single, optimized node:
\begin{equation}
    n_t^{*} = p_{\theta}(n_t^{1}, n_t^{2}, \ldots, n_t^{m}, \{R_{\text{selected}}\}) .
\end{equation}

Here the role of \( p_{\theta} \) is to ensure that information integration utilizes the content of documents correctly while keeping the backbones of the reasoning sequence. In this way, we enhance the depth and breadth of the node's information, and the optimized node \( n_t^{*} \) becomes more \underline{\textit{reliable and comprehensive}} in subsequent reasoning and decision-making processes. For all nodes generated in this layer, \( n_t^{*} \) represents the most valuable solution that LLM finds in the search space.

After generating the refined node \( n_t^{*} \), we combine \( n_t^{*} \) with the original prompt \( x \) to be the new prompt of task \(\mathcal{A}\) , and the model starts a new round of generation. This iterative process continues until the maximum number of iterations \( T \) is reached. At the end of the iterations, depending on the specific requirements of task \( \mathcal{A} \), the model output either the node \( n_T^{*} \) or the complete inference of \( n_T^{*} \) as decisions and inference.



\subsection{Discussions}
We further discuss why our method achieves the goals mentioned in the Introduction. First, to address the local challenge of correcting factual errors and avoiding the accumulation of errors, we need to perform fact-checking and corrections in a timely and continuous manner. Our \model method utilizes RAG technology to help LLMs efficiently and rapidly access information from relevant documents in an external library. The LLM integrates external knowledge to dynamically correct the factual errors caused by its possible limited knowledge, outdated information, or hallucinations within the reasoning context. Second, to address the global challenge of strategy optimization and searching for the optimal solution, we need to comprehensively evaluate the overall generating and searching strategy. Our \model method integrates factual accuracy and strategic feasibility to perform correction and optimization on each generated thought. In this way, the \model identifies and navigates the most promising branch within the search space. Finally, to reduce the risk of hallucinations, we need to make corrections during the generation process rather than after the entire reasoning process is completed. This is because the errors that develop and spread through the structure are difficult to entirely correct afterward, as they may have already polluted and magnified the whole generation. Thus, we design an online, incremental generation and correction tree structure. The experimental details in the next section demonstrate the effectiveness of our \model's capability in preventing the spread of errors and reducing hallucinations.

\begin{algorithm}
\caption{Retrieval Augmented Thought Tree (RATT)}
\begin{algorithmic}[1]
\State \textbf{Input:} Task prompt $x$, Node number $m$, Iteration time $T$, LLM $p_{\theta}$
\State $n_{t} \leftarrow \text{''''}$ \Comment{Initialize the node as an empty string}
\For{$t = 1$ \textbf{to} $T$} 
    \For{$l = 1$ \textbf{to} $m$} 
        \State $n_t^{(l)} \leftarrow p_{\theta}(\cdot \mid x, n_t)$ \Comment{Generate thoughts based on node $n_t$ and prompt $x$}

        \State $\{q_x,q_{n_t}^{(l)}\} \leftarrow \text{embed}(\{n_t^{(l)},x\}) $ \Comment{Embed the node and prompt}

        \State $q_t^{(l)} \leftarrow \text{CONCAT}(q_x, q_{n_{t}}^{(l)})$ \Comment{Concatenate the question and answer into a query vector}

        \State $\mathcal{R}_{\text{selected}} \leftarrow \text{RetrieveFromLibrary}(q_t^{(l)})$\Comment{Retrieve documents $\mathcal{R}_{\text{selected}}$ from Library}

        \State $n_t^{(l)*} \leftarrow p_{\theta}(\cdot \mid n_t^{(l)}, \mathcal{R}_{\text{selected}})$\Comment{Generate a refined node}

        \State $n_t \leftarrow \text{CONCAT}(n_t, n_t^{(l)*})$\Comment{Append the next refined node}    
    \EndFor
    \State $n_t^{*} \leftarrow p_{\theta}(\cdot \mid x, n_t)$\Comment{Refine and enhance the node}
\EndFor
\State $n_T^{*} \leftarrow p_{\theta}(\cdot \mid x, n_t^{*})$
\State \textbf{return} $n_T^{*}$ \Comment{Output $n_T^{*}$ as the final generation}
\end{algorithmic}
\end{algorithm}

\section{Experiments}
\subsection{Experimental Setup}
In this section, we propose four particularly challenging or representative tasks for LLM performance evaluation, and demonstrate the effectiveness of our \model approach. These multi-view tasks provide standard benchmarks to evaluate and compare the performance of our approach and baseline methods. Then we conduct an analysis of the results and discuss the strengths of our approach shown in the performance enhancement.

\textbf{Task Description.} We test our \model and baseline methods across four distinct tasks, each evaluating different aspects of the quality of LLM response. The \textit{Code Generation} and \textit{Creative Writing} are two comprehensive tasks, which \textit{Game of 24} focus on testing the logical and numerical reasoning capabilities. We also process a standard \textit{Hallucination Detection} to demonstrate the \model's ability in reducing hallucinations.

\textbf{Baselines Algorithms and Environmental Settings.} For baselines, we compare our approach with several prompting methods, including IO, CoT~\cite{wei2022chain}, CoT-SC~\cite{wang2022self}, ToT~\cite{TreeOfThoughts}, and RAT~\cite{RAT}. For the external library and database, we employ the \texttt{codeparrot/github-jupyter}~\cite{CodeParrot}~\footnote{\url{https://huggingface.co/datasets/codeparrot/github-jupyter}} as the search vector library of task \textit{Code Generation}, and the \texttt{English Wikipedia}~\cite{WikipediaMainPage} ~\footnote{\url{https://en.wikipedia.org}} as library of task \textit{Creative Writing} and \textit{Hallucination Detection}. We do not employ any external library for task \textit{Game of 24}, as the goal of this task is to examine the numerical reasoning ability of \model's thought tree structure. For the language model, we perform all the experiments on OpenAI API, GPT-3.5 Turbo~\cite{GPT35TurboFineTuning} model through the OpenAI platform~\footnote{\url{https://openai.com/index/openai-api/}}. For the implements, we perform the tasks on NVIDIA 4090.

\subsection{Code Generation}

\begin{figure}[htbp]
    \centering
    \includegraphics[width=1\linewidth]{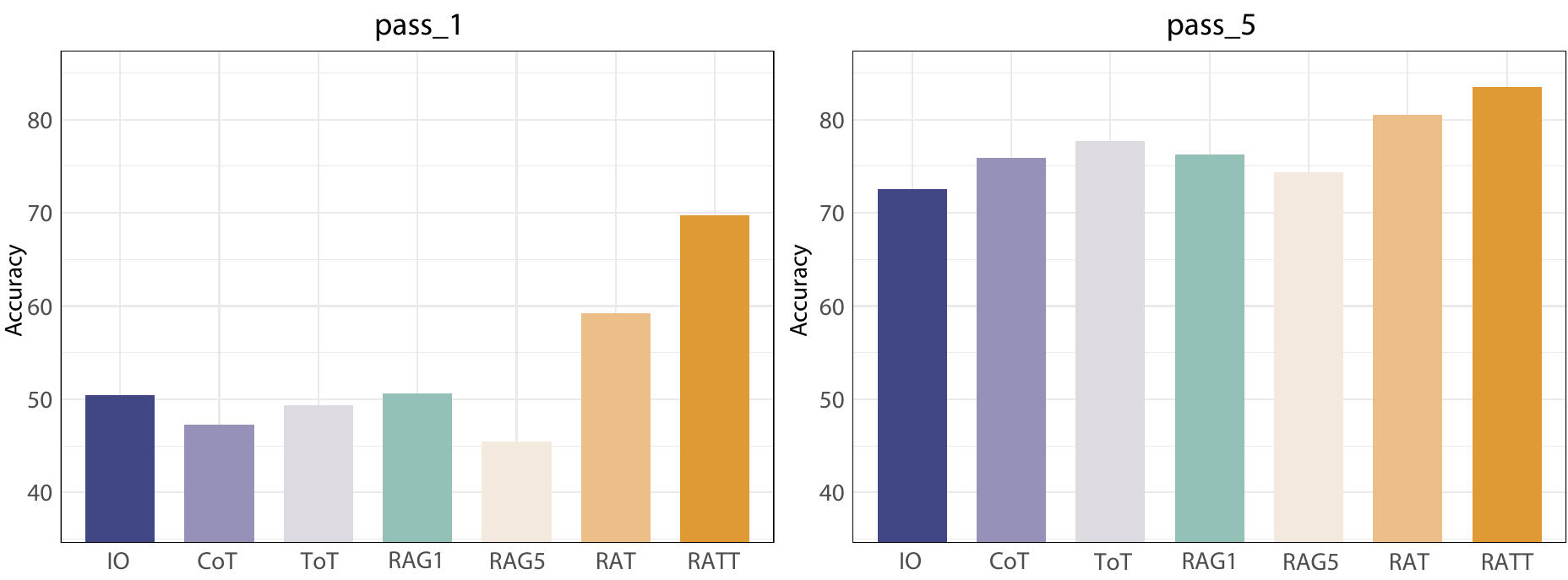}
    \caption{Comparison of code generation performance on \texttt{HumanEval} of different methods.}
    \label{fig:code}
\end{figure}

Code generation is a task where the model is tasked to understand programming prompts and produce functional and correct code based on those prompts. This task requires precise comprehension and application of programming logic, algorithms, and language syntax of the LLM. We conduct the code generation evaluation in \texttt{HumanEval}. \texttt{HumanEval} is a programming task benchmark specifically for the evaluation of coding capabilities of code generation models. In this task, we first construct a local dataset with \texttt{CodeParrot} and segment long texts into multiple chunks. Then we transform each chunk into embeddings with embedding model \texttt{text-embedding-ada-002}~\cite{OpenAI2024}~\footnote{\url{https://openai.com/index/new-and-improved-embedding-model/}}. After introducing tasks from \texttt{HumanEval}, the \model generates multiple potential answers. Finally, we validate the effectiveness of our \model approach using the original evaluation script from the \texttt{HumanEval GitHub} repository by \textit{pass@k} method~\cite{liu2024your}. This \textit{pass@k} method tests if there is at least one correct answer within \(k\) generated responses to assess the efficacy and accuracy of the model.
Our experiments on \texttt{HumanEval} have shown the highest performance improvement of \(38.05\%\) for \textit{pass@1} and \(15.12\%\) for \textit{pass@5} relatively, as summarized in Table~\ref{tab:code} and Figure~\ref{fig:code}.




\begin{table}[htbp]
    \centering
    \caption{Comparison of code generation performance on \texttt{HumanEval} of different methods.}
    \begin{tabular}{lcc}
        \toprule
        \textbf{Method} & \textbf{pass@1} & \textbf{pass@5} \\ \hline
        IO         & 50.49\%         & 72.56\%         \\
        CoT        & 47.31\%         & 75.88\%         \\
        ToT        & 49.36\%         & 77.73\%         \\
        RAG\_1 shot & 50.61\%         & 76.22\%         \\
        RAG\_5 shot & 45.49\%         & 74.39\%         \\
        RAT        & 59.27\%         & 80.49\%         \\
        RATT       & 69.70\%         & 83.53\%         \\ \bottomrule
    \end{tabular}
    \label{tab:code}
\end{table}

\subsection{Creative Writing}
Creative writing is a task that challenges LLMs to generate imaginative, coherent, and contextually rich text based on a variety of prompts. This task measures the LLMs' reasoning capabilities to innovate and think logically and coherently while enriching the context information. To further evaluate the \model's comprehensive performance on LLMs, we compare it with IO, CoT, ToT and RAT prompting. To assist this process, we utilize the \texttt{English Wikipedia}~\cite{WikipediaMainPage} library by searching this vast and diverse dataset to identify and retrieve the top few web pages that are most relevant to the given writing prompts. 

To ensure the consistency and integrity of the evaluation, we employ GPT-4~\cite{GPT4,naismith2023automated} to assist in assessing the quality of model outputs as \(50\%\) of the reviewers of this task while other reviewers are human annotation experts. Our scoring criteria encompass \textit{Soundness (Sound)}, \textit{Information Relevance (Rel)}, \textit{Content Coherence of Reasoning (Corr)}, and \textit{Clarity of Expression (Expr)}, which are uniformly applied. Each dimension receives a score up to 10, with increments of 0.5. The overall score is derived from the average of four metrics. 

\begin{figure}[h]
  \centering
  \includegraphics[width=1\linewidth]{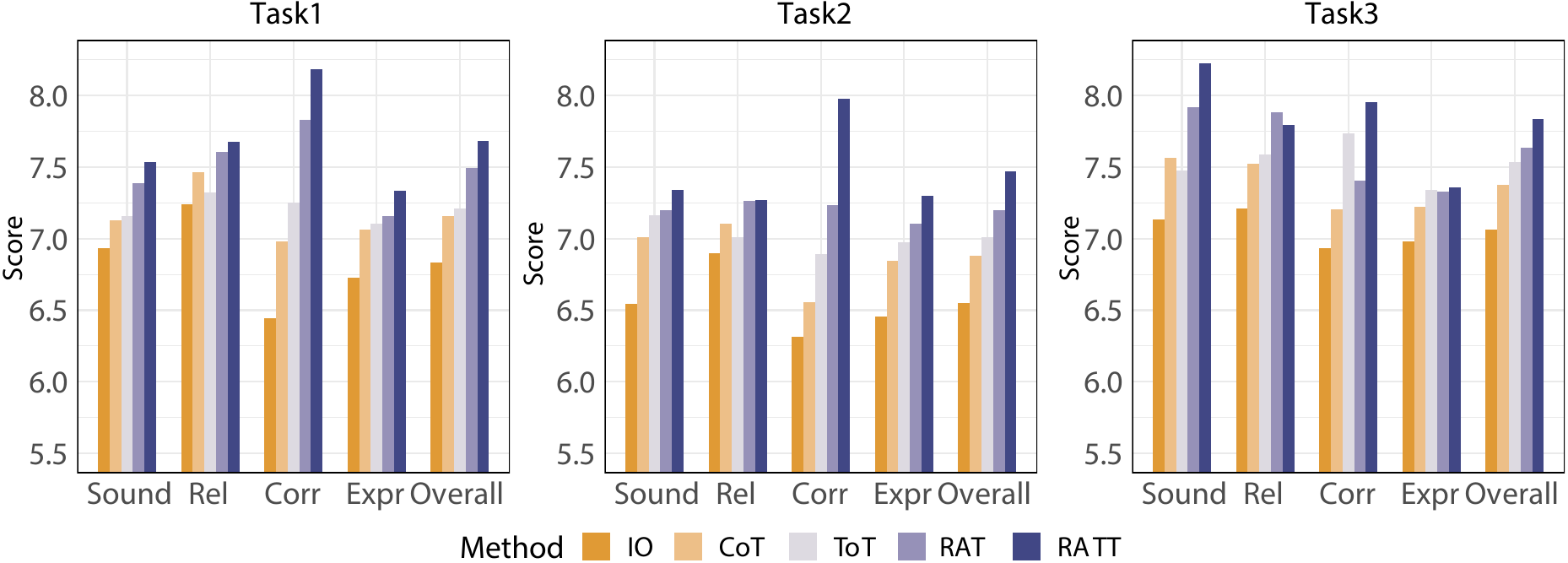}
  \caption{Comparison of different methods in task Creative Writing.}
  \label{fig:writing}
\end{figure}
\begin{figure}[h]
  \centering
  \includegraphics[width=1\linewidth]{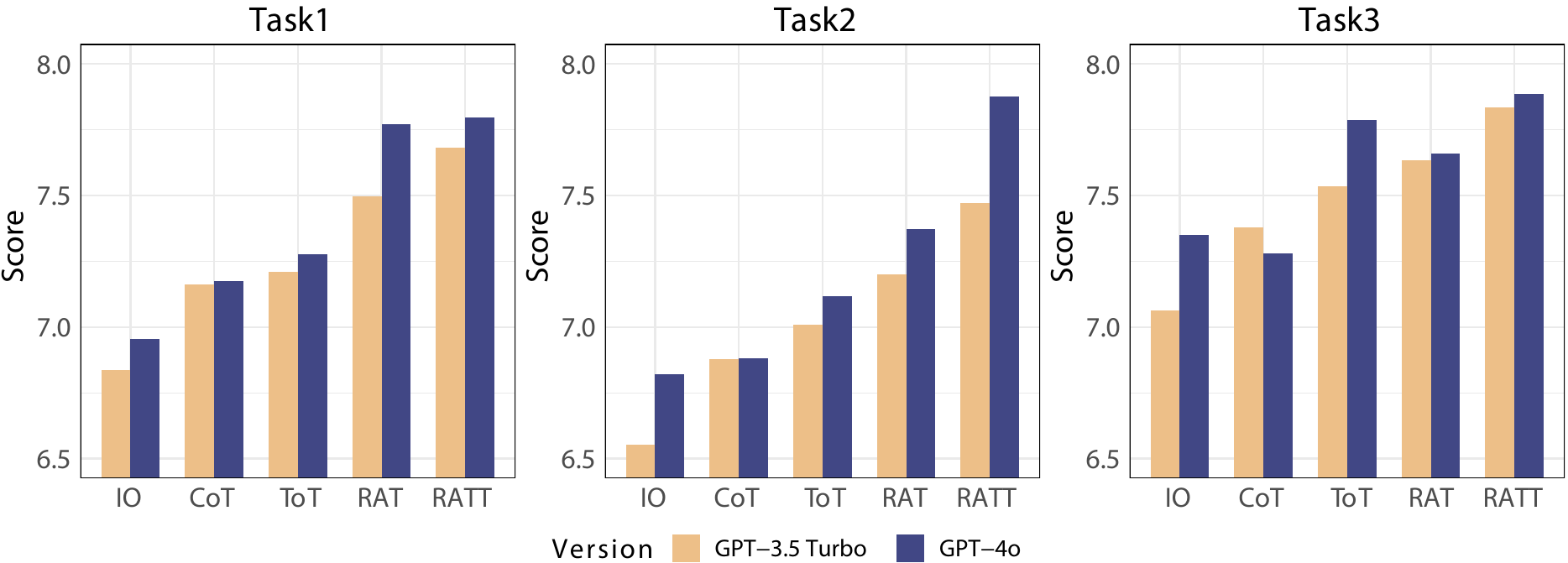}
  \caption{Improvement by different methods on GPT models in task Creative Writing.}
  \label{fig:34o}
\end{figure}

\begin{figure*}[h]
  \centering
  \includegraphics[width=0.99\linewidth]{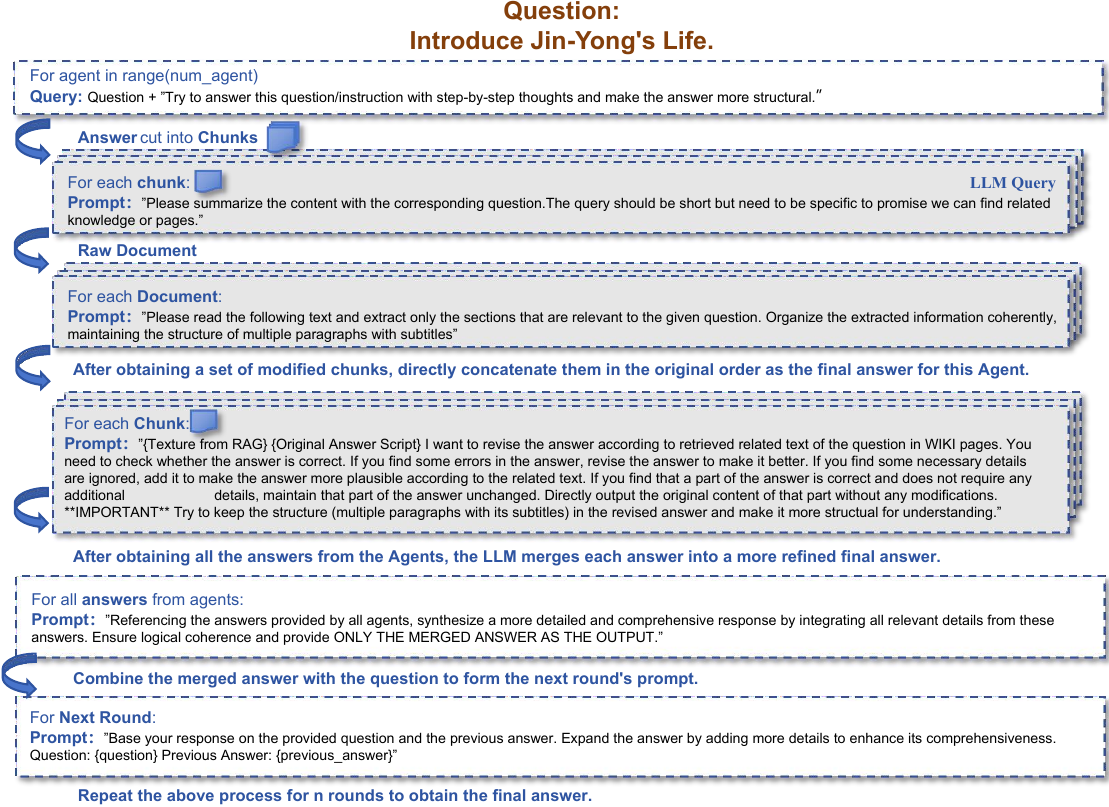}
  \caption{Prompting case of task Creative Writing.}
  \label{fig:prompt}
\end{figure*}

Here, we first conduct a comparison of various methods on GPT-3.5 Turbo to assess the performance across different metrics. As shown in Figure~\ref{fig:writing}, our approach demonstrates significant advantages over baseline methods in all dimensions and composite scores. We further study the improvements our method brings to both the GPT-3.5 Turbo and GPT-4o~\cite{HelloGPT4} models. As shown in Figure~\ref{fig:34o}, our approach and RAT significantly outperform the GPT-4o model. The results demonstrate the effectiveness of employing RAG on LLM in text generation and boosting the creative output of LLMs. In Figure~\ref{fig:prompt}, we present the detailed prompting and guidelines of the creative writing task.

\subsection{Hallucination Detection}
Following the method introduced in~\cite{ding2024retrieve}, we perform the standard hallucination detection. Hallucination detection refers to the process of identifying and assessing instances where language models generate text that is not grounded in reality~\cite{luo2024hallucination}. This task is crucial for evaluating the reliability of LLMs' text generation, as we target generating content that is accurate and trustworthy~\cite{gao2023retrieval,rawte2023survey}.

In this task, we utilized the TruthfulQA~\cite{lin2021truthfulqa} dataset to measure the truthfulness of outputs. Each ``correct answer'' in the dataset aims to reflect the truthfulness of a response to a given query. Our model generates answers to these queries, and then we measure the truthfulness of these answers in two ways:

\begin{figure}[h]
  \centering
  \includegraphics[width=1\linewidth]{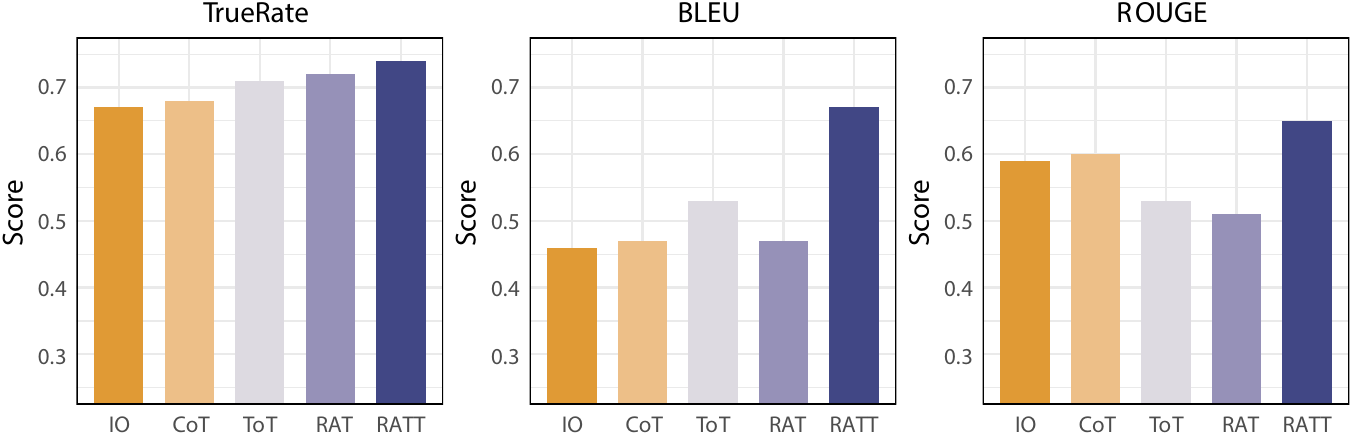}
  \caption{Comparison of hallucination metrics of different methods.}
  \label{fig:hallu}
\end{figure}

First, we directly calculate the similarity between the generated answers and the correct answers using BLEU and ROUGE scores. Here BLEU and ROUGE scores are the metrics designed to evaluate the quality of text by comparing machine-generated outputs against human references. Then we employ GPT-judge in~\cite{TruthfulQA} to predict the truthfulness of human-like answers in an end-to-end manner as part of the TruthfulQA project.



\subsection{Game of 24}

The ``Game of 24'' is a mathematical puzzle task that challenges an LLM's numerical reasoning capabilities. The game involves using the combination of addition, subtraction, multiplication, and division to manipulate four integers between 1 and 9 to arrive at the final result of 24. The rules are straightforward: players must use all four numbers exactly once, and the order of operations can vary as needed to solve the puzzle. This task is a benchmark for LLM mathematical reasoning~\cite{kim2023llm,ding2023everything}, as it requires arithmetic operations and strategic planning abilities to explore various combinations and operation sequences for the optimal solution.

\begin{table}[htbp] %
\centering
\small
\setlength{\tabcolsep}{8pt}
\caption{Success rates on Game of 24.}
\begin{tabular}{lc}
    \toprule
    Method       & Success (\%) \\
    \hline
    IO \rule{0pt}{10pt}          & 7.6 \\
    CoT          & 5.1 \\
    CoT-SC       & 11.2 \\
    ToT          & 14.0 \\
    RAT          & 9.0 \\
    RATT-3       & 25.9 \\
    RATT         & 31.8 \\
    \midrule
    Improvement  & 24.2 \\
    \bottomrule
\end{tabular}
\label{tab:methods_success}
\end{table}

We test the models on 100 randomly selected games from 4nums~\footnote{\url{https://www.4nums.com/game/difficulties/}}. For every test, if the model produces a valid equation that results in \(24\), we mark it as a success. From Table~\ref{tab:methods_success}, we can see that the \model achieves the highest success rate of \(31.8\%\), which significantly outperforms the GPT-3.5 Turbo model by \(24.2\%\) of improvement. Here the success of \model-3 is a success answer of \(24\) in \(3\) steps of generation, and \model in \(5\). IO, CoT, and CoT-SC can hardly help the model find a solution, while RAT can hardly bring advantages for LLM to win. These results underline our method's reasoning capability to enhance the numerical reasoning ability.

\section{Conclusion}
In this paper, we introduce \model, a novel thought tree structure that integrates both factual correctness and strategic coherence at every step of the reasoning process. This structure significantly enhances LLMs' logical coherence and decision-making efficiency in complex reasoning tasks by solving the existing problem of thought structures logically and globally. Our extensive experiments across various tasks demonstrate the effectiveness and superiority over existing methods, especially in improving factual correctness and logical coherence. By adopting \model on LLMs, we extend the boundaries of abilities that LLM performs in handling complex tasks. For future work, we plan to extend this LLM thought structure paradigm across various reasoning and LLM alignment tasks, and to conduct a thorough investigation into the generation strategies that our approach applies to the thought structure and solution space to study further insights into its underlying mechanisms and impacts.

While \model shows significant advancements and wide adaptability, there are several limitations that require further exploration, including high computational demands and limited scalability with very complex tasks or large external libraries. Second, the effectiveness of the reasoning process and final output heavily relies on the quality of the task prompt and the external library, which can affect the model's performance in scenarios with poorly organized prompts or limited external knowledge. Finally, adopting \model in tasks with unique requirements may have inherent limitations.

\section{Acknowledgments}
The author Xiting Wang was supported by the National Natural Science Foundation of China (NSFC) (NO. 62476279), Major Innovation \& Planning Interdisciplinary Platform for the ``Double-First Class'' Initiative, Renmin University of China, and the Fundamental Research Funds for the Central Universities, and the Research Funds of Renmin University of China No. 24XNKJ18.
\bibliography{aaai25}

\end{document}